\documentclass[final]{cvpr}

\usepackage{times}
\usepackage{epsfig}
\usepackage{graphicx}
\usepackage{amsmath}
\usepackage{amssymb}
\usepackage{float}
\usepackage{bbding}

\usepackage[pagebackref=true,breaklinks=true,colorlinks,bookmarks=false]{hyperref}

\def\cvprPaperID{****} 
\def\confYear{CVPR 2021}

\begin{document}

\title{End-to-end Generative Floor-plan and Layout with Attributes and Relation Graph}

\author{Xinhan Di \Envelope $^{1}$\\
IHome Company Nanjing\\
{\tt\small deepearthgo@gmail.com}
\and
Pengqian Yu $^{2}$\\
IBM Research Singapore\\
{\tt\small peng.qian.yu@ibm.com}
\and
Danfeng Yang $^{3}$\\
Radio Company Nanjing\\
{\tt\small breezeydf@gmail.com}
\and
Hong Zhu $^{4}$\\
IHome Company Nanjing\\
{\tt\small jszh0825@gmail.com}
\and
Changyu Sun $^{5}$\\
Huazhong University of Science and Technology\\
{\tt\small m202072256@hust.edu.cn}
\and
YinDong Liu $^{6}$\\
Tongji University\\
{\tt\small 2030237@tongji.edu.cn}
}

\maketitle

\begin{abstract}
In this paper, we propose an end-to-end model for producing furniture layout for interior scene synthesis from a random vector. This proposed model is aimed to support professional interior designers to produce interior decoration solutions more quickly. The proposed model combines a conditional floor-plan module of the room, a conditional graphical floor-plan module of the room, and a conditional layout module. Compared with the prior work on scene synthesis, our proposed three modules enhance the ability of auto-layout generation given the dimensional category of the room. We conduct our experiments on a proposed real-world interior layout dataset that contains $191,208$ designs from the professional designers. Our numerical results demonstrate that the proposed model yields higher-quality layouts in comparison with the state-of-art model. The dataset and codes are available at \url{https://github.com/CODE-SUBMIT/dataset3}.
\end{abstract}

\section{Introduction}

People spend lots of time indoors such as the bedroom, living room, office, gym and so on. Function, beauty, cost, and comfort are the keys to the redecoration of indoor scenes. The proprietor prefers demonstration of the layout of indoor scenes in several minutes nowadays. Therefore, online virtual interior tools become useful to help people design indoor spaces. These tools are faster, cheaper, and more flexible than real redecoration in real-world scenes. This fast demonstration is often based on the auto layout of indoor furniture and a good graphics engine. Machine learning researchers make use of virtual tools to train data-hungry models for the auto layout \cite{Dai_2018_CVPR,Gordon_2018_CVPR}. The models reduce the time of layout of furniture from hours to minutes and support the fast demonstration. 

Generative models of indoor scenes are valuable for the auto layout of the furniture. This problem of indoor scenes synthesis is studied since the last decade. One family of the approach is object-oriented which the objects in the space are represented explicitly \cite{10.1145/2366145.2366154,10.1145/3303766,Qi_2018_CVPR}. The other family of models is space-oriented which space is treated as a first-class entity and each point in space is occupied through the modeling \cite{10.1145/3197517.3201362}.

\begin{figure*}
\centering
\includegraphics[height=8.0cm]{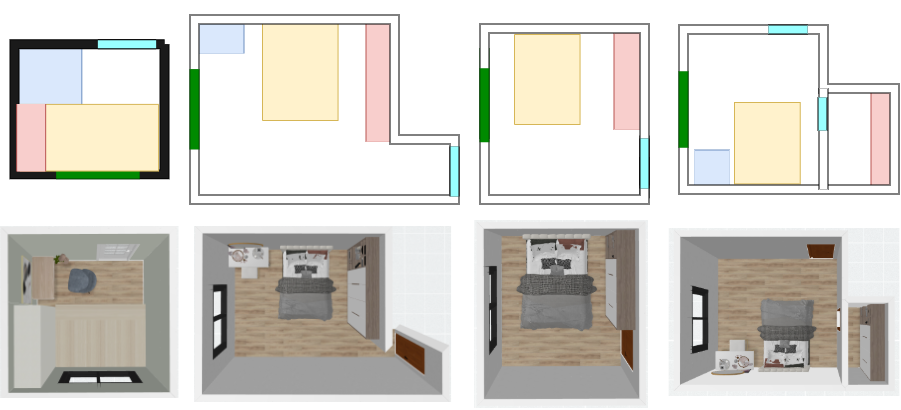}
\caption{Samples of the layout of bedroom with different dimensional size. The layouts are different for different dimensional sizes.}
\label{fig1}
\end{figure*}

Deep generative models are used for efficient generation of indoor scenes for auto-layout recently. These deep models further reduce the time from minutes to seconds. The variety of the generative layout is also increased. The deep generative models directly produce the layout of the furniture given an empty room. However, in the real world, the direction of a room is diverse in the real world. The south, north, or northwest directions are equally possible. The layout for the real indoor scenes is required to meet with a different dimensional size of an empty room as illustrated in Figure \ref{fig1}. 

Motivated by the above-mentioned challenge, we propose a model for producing the layout of the indoor scenes according to the dimensional category of a room in this paper. The model yields a design of layout of furniture according to the different dimensional size for a single type of room. This proposed model aims to support the interior designers to produce decoration solutions in the industrial process. In particular, this proposed adversarial model consists of several modules including a conditional floor-plan module of the room, a conditional graphical floor-plan module of the room, and a conditional layout module.

This paper is organized as follows: the related work is discussed in Section 2. Section 3 introduces the problem formulation. The methods of the proposed model are in Section 4. The proposed dataset is presented in Section 5. The experiments and comparisons with the state-of-art generative models can be found in Section 6. The paper is concluded with discussions in Section 7.

\section{Related Work}

Our work is related to data-hungry methods for synthesizing indoor scenes through the layout of furniture unconditionally or partially conditionally. 

\subsection{Structured data representation}
Representation of scenes as a graph is an elegant methodology since the layout of furniture for indoor scenes is highly structured. In the graph, semantic relationships are encoded as edges, and objects are encoded as nodes. A small dataset of annotated scene hierarchies is learned as a grammar for the prediction of hierarchical indoor scenes \cite{10.1145/3197517.3201362}. Then, the generation of scene graphs from images is applied, including using a scene graph for image retrieval \cite{Johnson_2015_CVPR} and generation of 2D images from an input scene graph \cite{Johnson_2018_CVPR}. However, the use of this family of structure representation is limited to a small dataset. In addition, it is not practical for the auto layout of furniture in the real world. 

\subsection{Indoor scene synthesis}
Early work in the scene modeling implemented kernels and graph walks to retrieve objects from a database \cite{Choi_2013_CVPR,Dasgupta_2016_CVPR}. The graphical models are employed to model the compatibility between furniture and input sketches of scenes \cite{10.1145/2461912.2461968}. However, these early methods are mostly limited by the size of the scene. It is therefore hard to produce a good-quality layout for large scene size. With the availability of large scene datasets including SUNCG \cite{Song_2017_CVPR}, more sophisticated learning methods are proposed as we review them below.

\subsection{Image CNN networks}
An image-based CNN network is proposed to encoded top-down views of input scenes, and then the encoded scenes are decoded for the prediction of object category and location \cite{10.1145/3197517.3201362}. A variational auto-encoder is applied to the generation of scenes with the representation of a matrix. In the matrix, each column is represented as an object with location and geometry attributes \cite{10.1145/3381866}. A semantically-enriched image-based representation is learned from the top-down views of the indoor scenes, and convolutional object placement a prior is trained \cite{10.1145/3197517.3201362}. However, this family of image CNN networks can not apply to the situation where the layout is different with different dimensional sizes for a single type of room.

\section{Problem Formulation}
\begin{figure*}
\centering
\includegraphics[height=7.5cm]{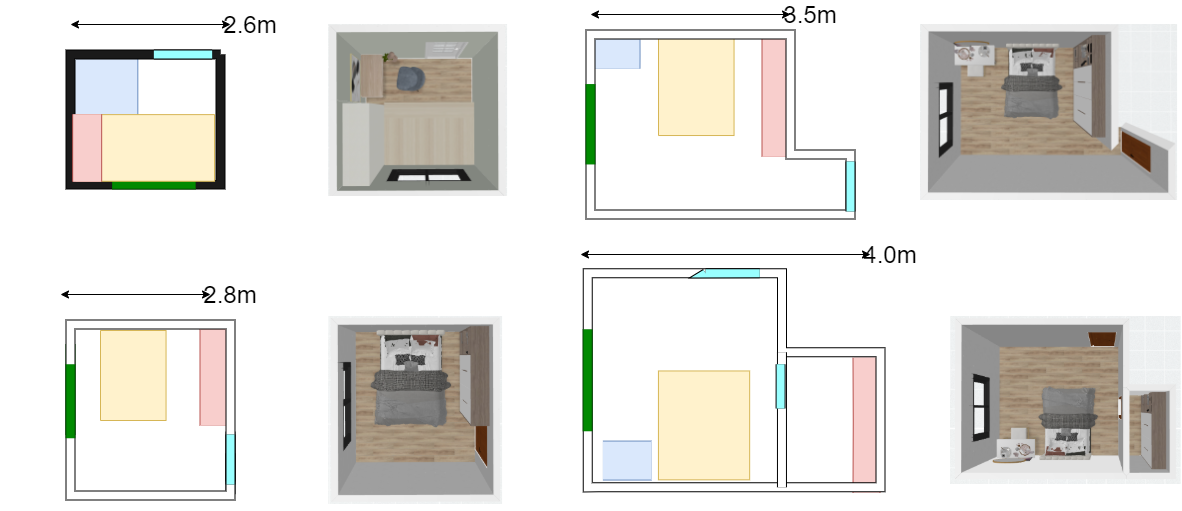}
\caption{If the length is less than 2.7 meters, then the layout is the type of tatami. If the length is between 2.7--3.4 meters, then the layout is the type of a common bedroom. If the length is more than 3.4 meters, the layout is the type of a common bedroom with other functionalities such as working or dressing.}
\label{fig2}
\end{figure*}

\subsection{Graph generative networks}
A significant number of methods have been proposed to model graphs as networks \cite{DBLP:journals/corr/abs-1709-05584,4700287}, the family for the representation of indoor scenes in the form of tree-structured scene graphs is studied. For example, Grains \cite{10.1145/3303766} consists of a recursive auto-encoder network for the graph generation and it is targeted to produce different relationships including surrounding and supporting. Similarly, a graph neural network is proposed for the scene synthesis. The edges are represented as spatial and semantic relationships of objects \cite{10.1145/3197517.3201362} in a dense graph. Both relationship graphs and instantiation are generated for the design of indoor scenes. The relationship graph helps to find symbolical objects and the high-lever pattern \cite{10.1145/3306346.3322941}.

\subsection{CNN generative networks}
The layout of indoor scenes is also explored as the problem of the generation of the layout. Geometric relations of different types of 2D elements of indoor scenes are modeled through the synthesis of layouts. This synthesis is trained through an adversarial network with self-attention modules \cite{DBLP:journals/corr/abs-1901-06767}. A variational autoencoder is proposed for the generation of stochastic scene layouts with a prior of a label for each scene \cite{Jyothi_2019_ICCV}. However, the generation of the layout is limited to produce a similar layout for a single type of room. 

\subsection{Problem definition}
We let a set of indoor scenes ${(x_{1},y_{1},l_{1}),\dots,(x_{N},y_{N},l_{N})}$ where $N$ is the number of the scenes, and $x_{i}$ is an empty indoor scene with basic elements including walls, doors and windows, $y_{i}$ is the corresponding layout of the furniture for $x_{i}$, and $l_{i}$ is the label corresponding to $x_{i}$.  Each $y_{i}$ contains the elements ${p_{j},s_{j}}$: $p_{j}$ is the position of the $j$-th element; $s_{j}$ is the size of the $j$-th element. Each element represents a furniture in an indoor scene $i$. We use $l$ to indicate the category of $x$ according to the area of a room. Figure \ref{fig2} illustrates an instance of the layout with different labels for a room: the left represents a small room with the layout of tatami, and the right represents a big room with the layout of common bedroom. 

\section{Methods}
\begin{figure*}
\centering
\includegraphics[height=8.5cm]{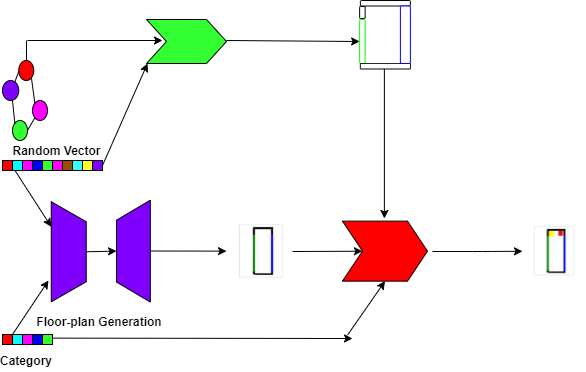}
\caption{Architecture of the proposed model which consists of a conditional generative module (in purple), a conditional graphical generative module (in green) and a conditional generative layout module (in red).}
\label{fig3}
\end{figure*}

In this section, we propose a model that produces the layout from a random vector and a given label. We define a model $M$ which produces the layout with a given random vector $z$ and a label $l$ in end-to-end training. The proposed model consists of the following modules. The first module is a conditional generative module that produces the floor-plan of a room with a given vector $z$ and a label $l$. The second module is a conditional graphical generative module that generates an abstract representation of the floor-plan when the input is $z$ and $l$. Thirdly, a conditional generative layout module is trained to produce the layout using the generative floor-plan, generative graphical floor-plan, and the label $l$. The proposed model as well as the modules are shown in Figure \ref{fig3}. In the following, we will discuss those modules as well as the training objectives.

\begin{figure}
\centering
\includegraphics[height=5cm]{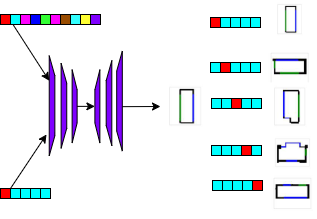}
\caption{Architecture of the proposed conditional generative module with the given random vector and labels of the dimensional size as input. It produces the floor-plan of an empty room with the corresponding dimension category.}
\label{fig4}
\end{figure}

\begin{figure}
\centering
\includegraphics[height=3.3cm]{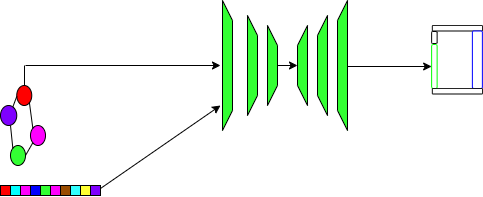}
\caption{Architecture of the proposed conditional graphical generative module with the given random vector and graphical representation of an empty room as input. It produces the structural representation of the empty room.}
\label{fig5}
\end{figure}

\subsection{Conditional generative module}
In the conditional generative model, the generation part $g_1$ takes the input of a random vector $z_{i}$ and $l_{i}$ and produces the floor-plan of a room. The discriminator part $d_{1}$ determines whether the generated floor-plan is real as illustrated in Figure \ref{fig4}. 

\subsection{Conditional graphical generative module}
In the conditional graphical generative module \cite{nauata2020house}, the generation part $g_{2}$ takes a random vector $z_{i}$ and the representation $r_{i}$ of the ground truth floor-plan of the room  as inputs. This representation $r_{i}$ encodes the number of walls, doors, windows, and the distance among them in an adjacency matrix. This is similar to the representation used in HouseGAN \cite{nauata2020house}. The discriminator part $d_{1}$ determines whether the generated box of the floor-plan is real as illustrated in Figure \ref{fig5}.

\begin{figure}
\centering
\includegraphics[height=4.2cm]{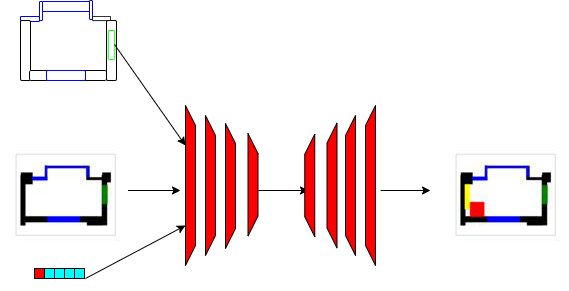}
\caption{Architecture of the conditional generative layout module with the generative floor-plan of an empty room, structural representation of an empty room, and the dimensional category as input. It produces the layout corresponding with the dimensional size of the room.}
\label{fig6}
\end{figure}

\subsection{Conditional generative layout module}
In the conditional generative layout module, the generation part $g_{3}$ takes the input of a generative floor-plan of a room, generated boxes of a floor-plan of a room, and the label $l$, produces a layout. The generator $g_{3}$ applies the floor-plan as the direct input, generated boxes as the graphical condition, and the label$l$ as the one-hot vector of the condition. It produces a layout which aims to generate a good layout plan according to different sizes of a room. The discriminator part $d_{3}$ determines whether the layout is real. This module is illustrated in Figure \ref{fig6}.  

\subsection{Training objectives}
We let $g_1$ denote the generator of the conditional generative module, $d_{1}$ denote its discriminator, and $g_{2}$ denote the generator of the conditional graphical generative module, $d_{2}$ denote its discriminator, $g_{3}$ denote the generator of the conditional generative layout module, and $d_{3}$ denote its discriminator. Given a random vector $z_{i}$ and a label $l_{i}$ which represents the category of the dimension of the room, the first generator $g_1$ is trained to produce a floor-plan of the room according to the given category $l_{i}$. The first discriminator $d_{1}$ is applied to determine whether the generated layout image is real. Similarly, the conditional graphical generator $g_{2}$ is trained with the given input, a random vector $z_{i}$, an encoded representation $r_{i}$ and category label $l_{i}$ to produce the abstract representation of the room $x_{i}$. Thirdly, the conditional layout generator $g_{3}$ is trained with the given input, a generative floor-plan $y_{g1}$, a generative graphical floor-plan $y_{g2}$ and the label $l_{i}$ to produce the layout according to the category $l_{i}$ of the room.

\subsubsection*{Conditional floor-plan generator training} 
To train the first generator network $g_{1}$,  the loss function $L_{g1}$ has the following form:
\begin{equation}
L_{G1} = L_{g1} + \lambda_{adv1}L_{adv1}
\end{equation}
where $L_{g}$ and $L_{adv1}$ denote the generation loss and the adversarial loss, respectively. Here $\lambda_{adv1}$ is the constant for balancing the multi-task training. In particular, $L_{g1}=M(g1(z_{i},l_{i}) - x_{1}^{gt})$, where $x_{1}^{gt}$ is the ground truth of floor-plan of a room and $M$ is a common distance function.

\subsubsection*{Conditional floor-plan discriminator training} 
To train the first discriminator network $d_{1}$, the first discriminator loss $L_{1D}$ has the following form
\begin{equation}
L_{1D} = -(1-y_{1n})\log(D1(P_{n1}^{0})) + y_{1n}\log(D1(P_{n1}^{1}))    
\end{equation}
where $y_{1n}=0$ if sample $P_{n}^{0}$ is drawn from the generator $g_{1}$, and $y_{1n}=1$ if the sample $P_{n}^{1}$ is from the ground truth. Here, $P_{n}^{0}$ denotes the rendered floor-plan image generated from the generator $g_{1}$, and $P_{n}^{1}$ denotes the rendered ground truth $x_{1}^{gt}$.

\subsubsection*{Conditional graphical floor-plan generator training} 
To train the second generator network $g_{2}$,  the loss function $L_{g2}$ has the following form:
\begin{equation}
L_{G2} = L_{g2} + \lambda_{adv2}L_{adv2}    
\end{equation}
where $L_{g2}$ and $L_{adv2}$ denote the generation loss and the adversarial loss, respectively. Here $\lambda_{adv2}$ is the constant for balancing the multi-task training. Paticularly, $L_{g2}=M(g2(z_{i},r_{i},l_{i}) - x_{2}^{gt})$, where $x_{2}^{gt}$ is the ground truth of structural representation of the room $x_{1}$ \cite{nauata2020house} and $M$ is a common distance function.

\subsubsection*{Conditional graphical floor-plan discriminator training} 
To train the second discriminator network $d_{2}$, the second discriminator loss $L_{2D}$ has the following form:
\begin{equation}
L_{2D} = -(1-y_{1n})\log(D2(P_{n1}^{0})) + y_{1n}\log(D2(P_{n1}^{1}))    
\end{equation}
where $y_{1n}=0$ if sample $P_{n}^{0}$ is drawn from the generator $g_{2}$, and $y_{1n}=1$ if the sample $P_{n}^{1}$ is from the ground truth. Here, $P_{n}^{0}$ denotes the rendered floor-plan representation generated from the generator $g_{2}$, and $P_{n}^{1}$ denotes the rendered ground truth $x_{2}^{gt}$.

\subsubsection*{Conditional layout generator training} 
To train the third generator network $g_{3}$,  the loss function $L_{g3}$ has the following form:
\begin{equation}
L_{G3} = L_{g3} + \lambda_{adv3}L_{adv3}    
\end{equation}
where $L_{g3}$ and $L_{adv3}$ denote the generation loss and the adversarial loss, respectively. Here $\lambda_{adv3}$ is the constant for balancing the multi-task training. Here, $L_{g3}=M(g3(g1(z_{i},l_{i}),g2(z_{i},l_{i},r_{i}),l_{i}) - y^{gt})$, where $y^{gt}$ is the ground truth of the layout according to the label $l_{i}$ for the room $x_{i}$ and $M$ is a common distance function.

\subsubsection*{Conditional graphical floor-plan discriminator training} 
To train the third discriminator network $d_{3}$, the third discriminator loss $L_{3D}$ has the following form:
\begin{equation}
L_{3D} = -(1-y_{1n})\log(D3(P_{n1}^{0})) + y_{1n}\log(D3(P_{n1}^{1}))    
\end{equation}
where $y_{1n}=0$ if sample $P_{n}^{0}$ is drawn from the generator $g_{3}$, and $y_{1n}=1$ if the sample $P_{n}^{1}$ is from the ground truth. Here, $P_{n}^{0}$ denotes the rendered layout generated from the generator $g_{3}$, and $P_{n}^{1}$ denotes the rendered ground truth $y^{gt}$.
\begin{figure*}[t]
\centering
\includegraphics[height=8.0cm]{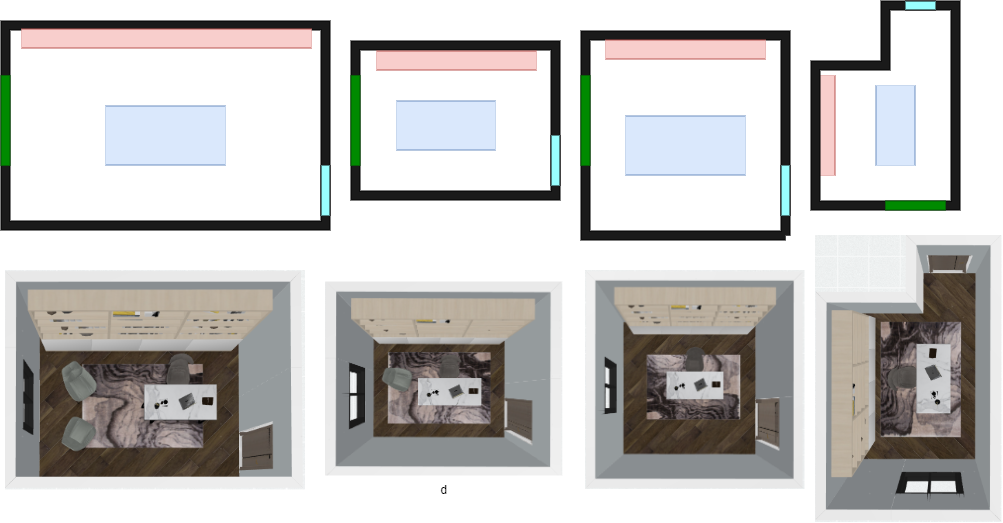}
\caption{The layout samples produced by our proposed model for $4$ different dimensional sizes of the study room.}
\label{fig8}
\end{figure*}

\begin{figure*}[t]
\centering
\includegraphics[height=7.0cm]{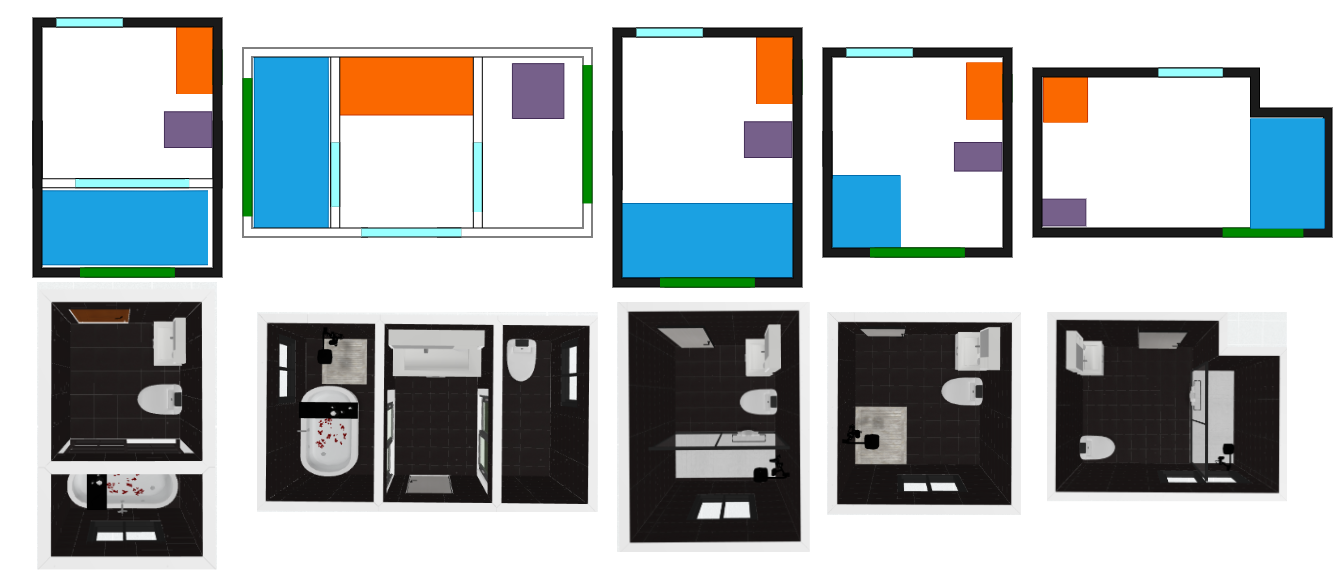}
\caption{The layout samples produced by our proposed model for $5$ different sizes of the bathroom.}
\label{fig9}
\end{figure*}
\section{Proposed dataset}

\begin{table*}[t]
\centering
\begin{tabular}{|p{2cm}|p{2cm}|p{2cm}|p{2cm}|p{2cm}|p{2cm}|p{2cm}|}
\hline
\multicolumn{1}{|c|}{}&\multicolumn{2}{|c|}{\text{Mode}}&\multicolumn{2}{|c|}{\text{IoU}}\\
\hline
\hfil Model    &\hfil PlanIT & \hfil Ours  &\hfil PlanIT &\hfil Ours \\
\hline
\hfil Bathroom   &$0.623\pm0.08$ &$0.721\pm 0.06$ &$0.629\pm0.09$ &$0.708\pm0.08$\\\hline
\hfil Bedroom   &$0.647\pm0.05$ &$0.719\pm 0.07$ &$0.612\pm0.06$ &$0.738\pm0.05$\\\hline
\hfil Study   &$0.602\pm0.06$ &$0.741\pm 0.05$ &$0.607\pm0.02$ &$0.729\pm0.07$\\
\hline
\end{tabular}
\caption{Evaluation Metrics for the Comparison with the state-of-art Model.}
\end{table*}\label{table1}
In this paper, we propose a dataset consist of indoor furniture layouts together with an end-to-end rendering image of the interior layout. This layout data is from designers at the real selling end where proprietors choose the design of the layout for their properties. Note that we labeled each layout according to the category of the dimension of the room as shown in Figure \ref{fig1}.  

\subsection{Interior layouts}
There are $60$ professional designers who work with an industry-lever virtual tool to produce a variety of designs. Among these designs, part of them are sold to the proprietors for their interior decorations. We collect these designs at the selling end and provide $191,208$ interior layouts. Each sample of the layout has the following representation including the categories of the furniture in a room, the position (x,y) of each furniture, the size (length, width, height) of each furniture, the position (x,y) of the doors and windows in the room, and the position (x,y) of each fragment of the walls, and the dimensional category of each room. Figure \ref{fig1} illustrates the samples of layouts adopted from the interior design industry. The dataset contains 3 main categories of rooms and 14 subcategories (dimensional category) of rooms. 

\subsection{Rendered layouts}
Each layout sample is corresponding to the rendered layout images. These images are the key demonstration of the interior decoration. These rendered images contain several views and we collect the top-down view as the rendered view as shown in Figure \ref{fig8}. Therefore, the dataset also contains $191,208$ rendered layouts in the top-down view. Each rendered layout is corresponding to a design. The rendered data is produced from an industry-lever virtual tool that has already provided missions of rendering layout solutions to the proprietors.

\section{Evaluation}
In this section, we present qualitative and quantitative results demonstrating the utility of our proposed model. Three main types of indoor rooms are evaluated including the bedroom, the bathroom, and the study room. $90\%$ of the $191,208$ samples are randomly chosen for training, and the remaining $10\%$ are used for the test. We compared the proposed model with the state-of-art models. 

\subsection{Evaluation metrics}
For the task of interior scene synthesis, we apply three metrics for the evaluation. Firstly, we use average mode accuracy for the evaluation. It is to measure the accuracy of the category of furniture for a layout corresponding with the ground truth. This average mode accuracy is defined as
\begin{equation}
\text{Mode} = \frac{\sum_{i=1}^{n}N_{i}^{1}}{\sum_{i=1}^{n}N_{i}^{total}} 
\end{equation}
where $N_{i}^{total}$ is the total number of $i$-th category of furniture in the ground truth dataset, and $N_{i}^{1}$ is the number of the $i$-th category of furniture in the generated layout in corresponding with the ground truth. For example, if the $i$-th furniture is in the predicted layout where the ground truth layout also contains this furniture, then it is calculated. Note that $n$ is the total number of the category of furniture.

Secondly, in order to evaluate the position accuracy of the furniture layout, we apply the Intersection over Union (IoU) between the predicted box of $i$-th furniture and the ground truth box.

We compare two baseline models for scene synthesis for the rooms. The results are shown in Figure \ref{fig8} and Figure \ref{fig9}. It can be observed that our model outperforms the state-of-art models in the following two aspects. First, the proposed model produces different layout according to the different dimensional categories of the main type of room. Moreover, our model predicts a good position of each furniture, while the state-of-art model sometimes predicts an unsatisfied position that is strongly against the knowledge of the professional interior designers. 

We also compare with the state-of-art model quantitatively. We report the performance metrics in Table 1. It can be seen that our model outperforms the state-of-art models in terms of the average mode accuracy and the position accuracy. 

\section{Discussion}
In this paper, we presented a model to predict the interior scene from a random vector. In addition, we propose an interior layouts dataset that all the designs are drawn from the professional designers. The proposed model achieves better performance than the state-of-art models on the interior layouts dataset. There are several avenues for future work. Our method is currently limited to the generation of layouts from a random vector.  It is also worthwhile to extend our work to generate layouts for multiple-rooms of larger sizes.  

{
\bibliographystyle{ieee_fullname}
\bibliography{egbib}
}

\end{document}